\documentclass[runningheads]{llncs}

\usepackage[T1]{fontenc}

\usepackage{graphicx}
\usepackage{comment}
\usepackage{amsmath,amssymb}
\usepackage{color}
\usepackage{url}
\usepackage{hyperref}
\usepackage{orcidlink}
\usepackage{wrapfig}
\usepackage{multirow}
\usepackage{booktabs}
\usepackage[normalem]{ulem}
\usepackage[utf8]{inputenc}
\usepackage{tcolorbox}
\usepackage{xcolor}
\newcommand{\xhdr}[1]{\textbf{#1}\:}

\newif\ifreview
\reviewfalse

\ifreview
	\usepackage{lineno}

	\linenumbers
\fi

\begin{document}

\def\SubNumber{000}

\def\GCPRTrack{Main Track}

\title{Investigating Structural Pruning and Recovery Techniques for Compressing Multimodal Large Language Models: An Empirical Study}

\ifreview
\else
	\titlerunning{Pruning and Recovery Techniques for Compressing MLLMs}

	\author{Yiran Huang\inst{1,2}\and
	Lukas Thede\inst{2,3} \and
	Massimiliano Mancini\inst{4}\and
Wenjia Xu\inst{5} \and \\
Zeynep Akata\inst{1,2}
    }
	
	\authorrunning{Huang et al.}
	
	\institute{Technical University of Munich, Germany \and Helmholtz Munich, Munich Center for Machine Learning, Germany \and University of Tübingen, Tübingen AI Center, Germany \and University of Trento, Italy\and Beijing University of Posts and Telecommunications, China\\
	\email{\{yiran.huang,zeynep.akata\}@tum.de}\\
    \email{Lukas.Thede@t-online.de}\\
    \email{massimiliano.mancini@unitn.it}\\
    \email{	xuwenjia@bupt.edu.cn}
    }
\fi

\maketitle              %

\begin{abstract}
While Multimodal Large Language Models (MLLMs) demonstrate impressive capabilities, their substantial computational and memory requirements pose significant barriers to practical deployment.
Current parameter reduction techniques primarily involve training MLLMs from Small Language Models (SLMs), but these methods offer limited flexibility and remain computationally intensive. 
To address this gap, we propose to directly compress existing MLLMs through structural pruning combined with efficient recovery training.
Specifically, we investigate two structural pruning paradigms---layerwise and widthwise pruning---applied to the language model backbone of MLLMs, alongside supervised finetuning and knowledge distillation. 
Additionally, we assess the feasibility of conducting recovery training with only a small fraction of the available data. 
Our results show that widthwise pruning generally maintains better performance in low-resource scenarios with limited computational resources or insufficient finetuning data. As for the recovery training, finetuning only the multimodal projector is sufficient at small compression levels (\textless 20\%). Furthermore, a combination of supervised finetuning and hidden-state distillation yields optimal recovery across various pruning levels. Notably, effective recovery can be achieved with as little as 5\% of the original training data, while retaining over 95\% of original performance. Through empirical study on two representative MLLMs, i.e., LLaVA-v1.5-7B and Bunny-v1.0-3B, this study offers actionable insights for practitioners aiming to compress MLLMs effectively without extensive computation resources or sufficient data.

\end{abstract}
\section{Introduction}
\begin{figure}[h]
\definecolor{prune}{RGB}{134, 163, 210}
\definecolor{recovery}{RGB}{112, 175, 131}
\vspace{-5pt}
    \centering
    \includegraphics[width=\linewidth]{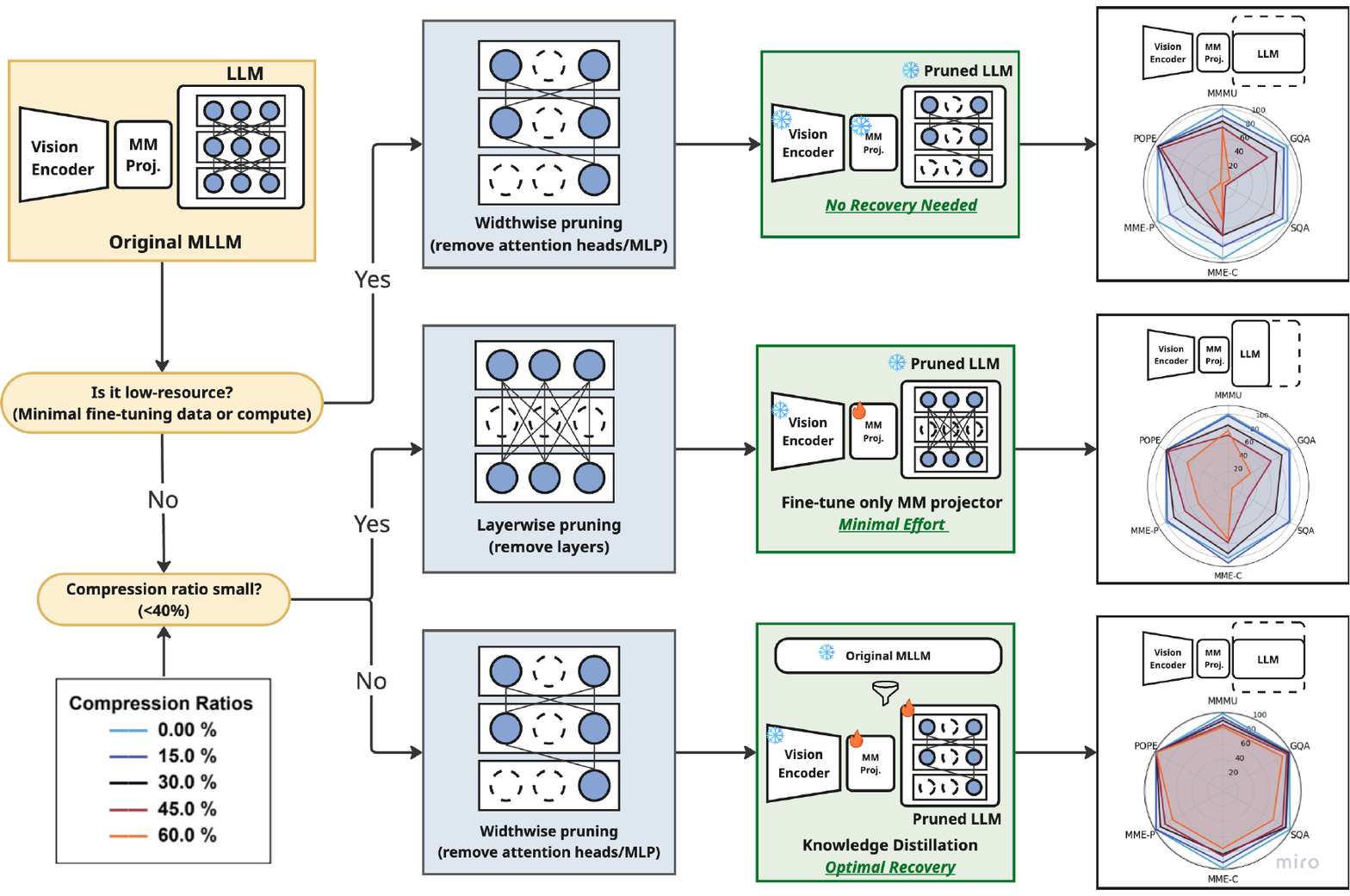}
    \vspace{-22pt}
    \caption{Compression decision flow for MLLMs. The left panel presents a decision flowchart that guides the choice of \textcolor{prune}{\textbf{pruning}} and \textcolor{recovery}{\textbf{recovery}} based on resource availability and compression ratio requirements.(i) widthwise pruning only (no recovery) in \textit{extremely low‐resource settings}; (ii) layerwise pruning with MM‐projector fine‐tuning for \textit{moderate compression} ($\leq 40\%$); and (iii) widthwise pruning + knowledge distillation for \textit{high compression} ($\geq 40\%$). The right panel shows spider plots of the retained performance across six multimodal benchmarks at 0--60 \% compression, demonstrating each strategy's effectiveness at various compression levels.}
    \label{fig:spiderweb_plot}
\vspace{-12pt}
\end{figure}

State-of-the-art MLLMs \cite{liu2023visualinstructiontuning,chu2023mobilevlm,chen2024internvl} based on Large Language Models (LLMs) \cite{touvron2023llama,jiang2024identifying} require substantial resources. 
For instance, models in the LLaVA \cite{liu2023visualinstructiontuning}  family commonly range from 7 billion up to 34 billion parameters, and even compact models like Bunny-v1.0 (3 billion parameters) present significant deployment challenges in resource-constrained environments.
Reducing the size of these models without compromising performance is crucial for adapting them to diverse deployment scenarios with varying resource constraints.

Existing approaches to this challenge focus mainly on building MLLMs from Small Language Models (SLMs) \cite{zhu2024comprehensive,he2024efficient,chu2023mobilevlm}. However, these methods suffer from fundamental limitations: they are constrained by the fixed size of the underlying SLM and require expensive training from scratch to meet target specifications.

We investigate an orthogonal and more flexible approach: structurally compressing the language components in MLLMs.
Specifically, we apply two pruning paradigms originally developed for LLMs to the MLLM setting. The first, layerwise pruning, removes entire transformer blocks, leveraging evidence that many layers are redundant \cite{fan2019reducing,sajjad2023effect}. The second, widthwise pruning, drops unimportant attention heads and MLP neurons, reflecting observations that only a subset of these sub-components is essential \cite{voita2019analyzing,michel2019sixteen,mccarley2019structured,hudson2019gqa}. Crucially, we pair these pruning strategies with various recovery training methods, including supervised finetuning and knowledge distillation on both logits and hidden states. Finally, we vary the pruning ratio and the amount of available data to map out the accuracy/efficiency frontier.
Our systematic empirical analysis provides insights into how pruning and recovery techniques impact MLLM performance under various compression levels and data availability scenarios. Specifically, we found:
\begin{itemize}
        \item Widthwise pruning is more effective in low-resource scenarios, i.e., when computational resources or sufficient finetuning data are unavailable. With recovery training, layerwise pruning is better for small ratios while widthwise pruning usually outperforms it at larger ones (\textgreater 40\%).
        \item Finetuning only the multimodal projector is sufficient at small compression levels (\textless 20\%), 
        as pruning has a minimal impact on the language model itself, but damages the multimodal alignment.
        \item Supervised finetuning with hidden-state distillation consistently provides the highest performance recovery across all compression ratios.
        \item Higher pruning ratios require larger amounts of data for effective recovery, while minimal data (5\%) can suffice at moderate compression levels(\textless 30\%).
\end{itemize}
We highlight our key findings in Figure \ref{fig:spiderweb_plot}. 
Our findings enable practitioners to efficiently compress MLLMs, allowing researchers to build upon empirically supported strategies without undertaking extensive experimentation themselves. 

\section{Related Work}
\label{related_work}

{\bf Pruning.}
\emph{Unstructured pruning}\cite{dong2017learningprunedeepneural,frankle2019lotterytickethypothesisfinding,lee2020signalpropagationperspectivepruning,park2020lookaheadfarsightedalternativemagnitudebased,sanh2020movementpruningadaptivesparsity,farina2024multiflow} removes individual weights or neurons. While such approaches can achieve strong compression rates with minimal accuracy trade-offs, they usually require specialized hardware or software for effective acceleration. 
In contrast, \emph{structured pruning} \cite{ding2019centripetalsgdpruningdeep,li2017pruningfiltersefficientconvnets,liu2021groupfisherpruningpractical,you2019gatedecoratorglobalfilter} eliminates entire groups of parameters to reduce both the model size and its computational overhead. 
Within LLMs, recent work demonstrates that structured pruning can remove full layers or attention heads with modest performance drop \cite{fang2023depgraph,ma2023llm,xia2024shearedllamaacceleratinglanguage}.
Dynamic schemes that adapt the pruning pattern during training have also been explored \cite{dery2024everybodyprunenowstructured}. Our study builds on these advances, concentrating on structured pruning for the language-model backbone of MLLMs and systematically pairing them with recovery training.

{\bf Other Compression Methods.} \emph{Quantization} \cite{bai2021binarybertpushinglimitbert,yao2022zeroquantefficientaffordableposttraining,Zafrir_2019} reduces parameter precision to shrink memory, while \emph{low-rank factorization} \cite{hsu2022languagemodelcompressionweighted,hu2021loralowrankadaptationlarge,lan2020albertlitebertselfsupervised,ashkboos2024slicegpt} approximates large weight matrices with low-rank products. These methods are complementary to pruning while do not directly address architectural redundancy. 
We focus on pruning to permit fine-grained control over model structure.

{\bf Knowledge distillation (KD)} transfers knowledge from a large teacher model to a smaller student model \cite{hinton2015distillingknowledgeneuralnetwork,Gou_2021,sanh2020distilbertdistilledversionbert}. 
For language models KD has been applied to classification \cite{sanh2020distilbertdistilledversionbert,liang2021mixkdefficientdistillationlargescale} and generation \cite{xu2024surveyknowledgedistillationlarge}, with extensions to hidden-state mimicry \cite{jiao2020tinybertdistillingbertnatural,sun2019patientknowledgedistillationbert}, attention alignment \cite{wang2020minilmdeepselfattentiondistillation}, and reverse-KL objectives \cite{gu2023knowledge}.
We adopt KD as a recovery mechanism after aggressive pruning and empirically compare its benefits with those of lighter finetuning schemes.

{\bf Efficient MLLMs.} 
Current efforts to build lightweight multimodal systems rely on SLMs such as Phi-2 in LLaVA-Phi \cite{zhu2024llava}, specialized projector designs in MobileVLM \cite{chu2023mobilevlm}, or careful data curation in Bunny-v1.0 \cite{he2024efficient}.
While effective, these approaches inherit the fixed size of the underlying SLM. Our study specifically addresses methods for customizing the size of existing MLLMs through structured pruning and recovery strategies.

\section{Methodology}
\label{Pruning Methodology}
\xhdr{Notation.} Given a triplet $\mathbf{X} = \{\mathbf{x}_v, \mathbf{x}_p, \mathbf{x}_r\}$, the objective of an MLLM $m_\theta$, parameterized by $\theta=\{\psi,\phi,\mathbf{W}\}$, is to generate a response $\mathbf{x}_r$ based on an input image $\mathbf{x}_v$ and a text prompt $\mathbf{x}_p$, such that $m_\theta(\mathbf{x}_v, \mathbf{x}_p) = \mathbf{x}_r$. The MLLM typically consists of a vision encoder $g_\psi(\cdot)$, an LLM $f_{\phi}(\cdot)$, and a multimodal projector $\mathbf{W}$ aligning the two modalities. The prompt $\mathbf{x}_p$ is tokenized into $\mathbf{T}_p$, while the vision encoder processes the image $\mathbf{x}_v$ to extract visual features, which are then converted into language embedding tokens $\mathbf{T}_v$ via the multimodal projector:
\begin{equation}
    \mathbf{T}_v=\mathbf{W} \cdot g_\psi(\mathbf{x}_v) \ \ \text{and} \ \ f_\phi(\mathbf{T}_v \odot \mathbf{T}_p) = \mathbf{x}_r.
\end{equation}

The concatenated visual tokens $\mathbf{T}_v$ and prompt tokens $\mathbf{T}_p$ are fed into the LLM’s $M$ layers, producing hidden states $\{{\mathbf{H}_i} \in \mathbb{R}^{T \times d}\}^M_{i=1}$, where $T$ is the number of tokens and $d$ is the hidden dimension. Finally, the probabilities $p_{m_\theta}(\mathbf{x}_r|\mathbf{x}_v, \mathbf{x}_p,\tau)$ are computed by passing the final hidden state through the classification head with softmax temperature $\tau$.

\subsection{Pruning}
Large Transformers are largely over-parameterized, as whole layers can be dropped with little accuracy loss \cite{fan2019reducing,sajjad2023effect}, and only a few attention heads or MLP units per layer truly matter \cite{voita2019analyzing,michel2019sixteen,mccarley2019structured,hudson2019gqa}.
Motivated by these findings, we explore two pruning paradigms specifically targeting the language model backbone within MLLMs: layerwise pruning, which removes entire transformer layers, and widthwise pruning, which eliminates the least important components within each layer.
To determine which layers or components to prune, we draw a small subset of $n$ samples from the original visual instruct-tuning dataset as the calibration dataset $\mathcal{D} = \{\mathbf{x}^j_v, \mathbf{x}^j_p, \mathbf{x}^j_r\}_{j=1}^{n}$. The importance of each layer or component is assessed, and those with the lowest importance are pruned.

\xhdr{Layerwise Pruning.}
\label{layer_prune}
To identify the redundant layers, we use the Block Influence (BI) score \cite{men2024shortgpt}, which quantifies the importance of layer $i$ through the cosine distance between input $\mathbf{H}_i$ and output hidden states $\mathbf{H}_{i+1}$. The key assumption is that layers that cause larger changes in hidden states have a greater influence on model performance. The BI score of layer $i$ is then calculated by
\begin{equation}
    \text{BI}_i(\mathcal{D}) = 1 - \mathbb{E}_{\mathbf{X} \sim \mathcal{D}, t}\left[\frac{\mathbf{H}^\mathsf{T}_{i,t} \mathbf{H}_{i+1,t}}{\|\mathbf{H}_{i,t}\|_2 \|\mathbf{H}_{i+1,t}\|_2}\right],
\end{equation}
where $\mathbf{H}_{i,t}$ represents the $t^{th}$ row of $\mathbf{H}_i$. After calculating the BI scores, the layers are ranked by importance, and those with the lowest scores are pruned. 

\xhdr{Widthwise Pruning.}
\label{width_prune}
To address the widthwise redundancy, we apply dependency-based structural pruning. Following \cite{fang2023depgraph} and \cite{ma2023llm}, we build a dependency graph inside each LLM layer. Let $N_{i}$ and $N_{j}$ represent two neurons in the layer, where $\text{In}(N_{i})$ and $\text{Out}(N_{i})$ represent the neurons connected to $N_{i}$ as inputs and outputs, respectively. Neuron $N_{j}$ is dependent on on $N_{i}$ if
\begin{equation}
    \displaystyle N_{j} \in \text{Out}(N_{i}) \cap \text{Num}_{\text{In}(N_{j})} = 1 \text{, or } \displaystyle N_{j} \in \text{In}(N_{i}) \cap \text{Num}_{\text{Out}(N_{j})} = 1,
\end{equation}
where $\text{Num}_{\text{In}(N_{j})}$ refers to the number of input neurons of $N_{j}$ and $\text{Num}_{\text{Out}(N_{j})}$ is the number of the output neurons of $N_{j}$. In words, $N_{i}$ is the only downstream or upstream node of $N_{j}$. If neuron $N_{i}$ is pruned, all its dependent neurons $N_{j}$ must also be pruned. This process results in a set of dependency graphs ${{G}} = \{w_i^k\}_{i=1}^M$, where $M$ is the number of structures in the graph and ${w_i^k}$ represents the $k^{th}$ weight parameter within a structure. We assess their importance at the group level since all weights within a graph must be pruned together. Group importance is evaluated by comparing the loss of vision language modeling ${\displaystyle \mathcal{L}_{CE}(m_\theta(\mathbf{x}_v,\mathbf{x}_q),\mathbf{x}_r)}$ in the calibration data set, with and without weight. To efficiently approximate the importance, we apply a Taylor expansion using gradient information:
\begin{equation}
    \displaystyle I_{w_i^k}(\mathbf{X}) = |\mathcal{L}_{CE}(\mathbf{X},m_\theta) - \mathcal{L}_{CE}(\mathbf{X},m^{w_i^k=0}_\theta)|  \approx \left| \frac{\partial \mathcal{L}_{CE}(\mathbf{X}, m_\theta)}{\partial w_i^k} w_i^k \right| \,.
\end{equation}

We then prune the graphs with the lowest group importance $I_{G}$:
 \begin{equation}
     \textstyle{I_{G}(\mathcal{D}) = \mathbb{E}_{\mathbf{X}\sim \mathcal{D}} \left[\sum_{i}^M\sum_{k}{I_{w_i^k}}(\mathbf{X})\right]} \,.
 \end{equation}

\subsection{Recovery Training}
\label{Retraining}
Pruning a large multimodal language model results in performance degradation, affecting both language modeling and cross-modality alignment. To mitigate this, we investigate two recovery training methods: supervised finetuning (Sec. \ref{Retraining_sft}) and knowledge distillation (Sec. \ref{retraining_kd}). We consider the original teacher model $m^\text{T}_\theta$, the pruned student model $m^\text{S}_{\theta^\prime}$, and a recovery dataset $\mathcal{D}$.

\vspace{-10pt}
\subsubsection{Recovery Training with Supervised Finetuning~(FT).}
\label{Retraining_sft}
We first focus on training only the multimodal projector to realign the vision and language spaces.
Second, we jointly finetune the projector and the pruned language model while keeping the vision encoder fixed, as finetuning the vision encoder does not improve performance \cite{karamcheti2024prismatic}.
We use the cross-entropy loss for supervised finetuning:
\begin{equation}
    \mathcal{L}_{sft}(m^\text{S}_{\theta^\prime}, \mathcal{D})=\mathbb{E}_{\mathbf{X}\sim\mathcal{D}}[\mathcal{L}_{CE}(m^\text{S}_{\theta^\prime}(\mathbf{x}_v,\mathbf{x}_p),\mathbf{x}_r)].
\end{equation}

\vspace{-10pt}
\subsubsection{Recovery Training with Knowledge Distillation.}
\label{retraining_kd}
KD
allows the pruned model to regain lost performance by mimicking the decision-making process of the more capable teacher (original model). 
We explore two main strategies, logits-based KD and hidden state based KD.

\xhdr{Logits-based KD} aligns the output probability distributions of the pruned model with those of the teacher model. The logits-based KD loss is defined as
\begin{equation}
    \mathcal{L}_{logits}(m^\text{S}_{\theta^\prime}, m^\text{T}_\theta, \mathcal{D}) =  \mathbb{E}_{\mathbf{X}\sim\mathcal{D}}\left[\mathcal{L}_{KD}(p_{m^\text{T}_\theta}(\mathbf{x}_r|\mathbf{x}_v, \mathbf{x}_p,\tau), p_{m^\text{S}_{\theta^\prime}}(\mathbf{x}_r|\mathbf{x}_v, \mathbf{x}_p,\tau))\right].
\end{equation}
We explore two losses to evaluate the differences between the logit distributions of the student $p_\theta$ and the teacher $q_{\theta^\prime}$: Kullback–Leibler divergence (KL), denoted as  $\mathbf{KL} (p_\theta \Vert q_{\theta^\prime})$ and its reversed form (RKL), denoted as ${\mathbf{KL}(q_{\theta^\prime} \Vert p_\theta)}$. The standard KD objective, minimizing the approximated forward KL, encourages the student distribution to match all modes of the teacher distribution.
In contrast, using RKL encourages $q_{\theta^\prime}$ to focus on the major modes of $p_\theta$ while assigning low probabilities to its less significant regions. This helps the student model avoid learning unnecessary long-tail variations of the teacher distribution and instead focus on generating more accurate responses \cite{gu2023knowledge,holtzman2019curious}.

\xhdr{Hidden State Matching} involves aligning the pruned model’s intermediate representations (hidden states) $\mathbf{H}^{m^\text{S}_{\theta^\prime}}_i$ with the teacher model's $\mathbf{H}^{m^\text{T}_\theta}_i$. The corresponding loss for a layer $i$ can be defined as
\begin{equation}
    \mathcal{L}_{match}(m^\text{S}_{\theta^\prime}, m^\text{T}_\theta, \mathcal{D}) = \mathbb{E}_{\mathbf{X}\sim\mathcal{D}}\left[\mathcal{L}_{feat}(\mathbf{H}^{m^\text{S}_{\theta^\prime}}_i,\mathbf{H}^{m^\text{T}_\theta}_i)\right],
\end{equation}
where $\mathcal{L}_{feat}$ refers to a feature matching loss. Both \cite{yang2024clip} and \cite{popp2024zero} suggest that applying a feature-based L2 distillation loss improves the student model’s performance, particularly for pre-trained vision-language models. Consequently, we employ L2 loss as the feature matching loss $\mathcal{L}_{feat} =\|\cdot - \cdot\|_2^2$. 
The total loss for recovery training is computed as:
\begin{equation*}
    \mathcal{L}(m^\text{S}_\theta, m^\text{T}_\theta, \mathcal{D}) = \alpha \mathcal{L}_{sft}(m^\text{S}_{\theta^\prime}, \mathcal{D}) + \beta \mathcal{L}_{logits}(m^\text{S}_{\theta^\prime}, m^\text{T}_\theta, \mathcal{D}) +  \gamma \mathcal{L}_{match}(m^\text{S}_{\theta^\prime}, m^\text{T}_\theta, \mathcal{D})
\end{equation*}
where $\alpha$, $\beta$, and $\gamma$ are the coefficients that balance three loss components.

\section{Experiments}
\label{Experiments and results}

\xhdr{Experimental Setup.} We evaluate pruning and recovery methods on both a large-scale MLLM model (LLaVA-v1.5-7B (LLaVA) \cite{liu2024improved}) and a smaller-scale MLLM model (Bunny-v1.0-3B (Bunny) \cite{he2024efficient}). 
We provide a detailed overview of the model architectures in Appendix~\ref{Model architecture of LLaVA and Bunny}. 
For both models, we exclusively use their visual instruction tuning datasets: LLaVA-v1-5-mix665k \cite{liu2024improved} for LLaVA and Bunny-695K \cite{he2024efficient} for Bunny. During pruning, we randomly select 10 samples from the training dataset as the calibration dataset to compute the importance. For recovery training, we experiment with various portions of the original dataset ~( 5\%, 10\%, 20\%, and 100\%) for recovery. We set the distillation temperature to 2.0 for logits-based distillation and use the final layer representation for hidden state matching (see Appendix \ref{L2 layer comparison}).
We evaluate the pruned and recovery-trained models on visual question-answering tasks using GQA \cite{hudson2019gqa} and SQA-I \cite{lu2022learn}, as well as instruction-following tasks with POPE \cite{li2023evaluating}, MME-Cognition, MME-Perception \cite{yin2023survey}, and MMMU \cite{yue2023mmmu}. To ensure consistency, we use the lmms-eval suite \cite{lmms_eval2024} for all evaluations. For clearer comparisons, we calculate the relative performance as a percentage of the original (uncompressed) model’s performance on each benchmark. 

\subsection{The effect of pruning on the model performance and resources usage}
\label{Experiments_pruning}
\begin{table}[t]
\caption{Pruning results for LLaVA-v1.5-7B and Bunny-v1-3B. Size is the number of total parameters of the model, while the compression ratio (Ratio) indicates the proportion of remaining language model parameters compared to the pre-pruning state. For both models, width-wise pruning results in better performance without finetuning compared to depth-wise pruning.}
\vspace{-10pt}
\begin{center}
\resizebox{\linewidth}{!}{
\begin{tabular}{lrrrrrrrrrr}
\toprule
\textbf{Method} & \textbf{Size} & \textbf{PruneRatio} & \textbf{MMMU} & \textbf{GQA} & \textbf{SQA} & \textbf{MME-C} & \textbf{MME-P} & \textbf{POPE} & \textbf{AVG} & \textbf{AVG-\%} \\
\midrule
LLaVA-v1.5-7B& 7.0B &  & 35.10 & 61.98 & 68.67 & 363.21 & 1511.33 & 86.99 & 62.28 & 100.00\% \\
\midrule
Width-wise & 6.3B & 15\% & 32.40 & 59.34 & 63.21 & 268.93 & 1432.47 & 86.57 & 57.79 & 92.79\% \\
& 5.5B & 30\% & 31.00 & 52.59 & 54.29 & 253.21 & 1174.93 & 86.29 & 52.43 & 84.17\% \\
& 4.8B & 45\% & 27.60 & 20.86 & 12.10 & 70.00 & 347.45 & 45.96 & 22.11 & 35.49\% \\
& 4.0B & 60\% & 23.30 & 0.43 & 0.40 & 2.14 & 19.24 & 3.94 & 4.88 & 7.84\% \\ 
\midrule
Depth-wise & 6.3B & 15\% & 31.80 & 42.77 & 55.23 & 202.14 & 701.83 & 86.38 & 46.09 & 74.00\% \\
& 5.5B & 30\% & 32.70 & 42.18 & 59.64 & 210.71 & 921.88 & 78.69 & 47.61 & 76.43\% \\
& 4.8B & 45\% & 26.90 & 14.39 & 3.82 & 132.86 & 616.63 & 51.69 & 24.04 & 38.60\% \\
& 4.0B & 60\% & 25.80 & 0.00 & 0.00 & 0.00 & 0.00 & 0.00 & 4.30 & 6.90\% \\
\midrule 
Bunny-v1\-0-3B& 3.2B &  & 34.10 & 54.72 & 70.70 & 289.30 & 1487.71 & 87.82 & 59.65 & 100.00\% \\
\midrule 
Width-wise & 2.8B & 15\% & 30.90 & 51.83 & 65.64 & 242.50 & 1207.85 & 87.94 & 54.50 & 95.48\% \\
& 2.5B & 30\% & 28.40 & 45.65 & 55.73 & 199.64 & 807.95 & 87.13 & 47.04 & 87.57\% \\
& 2.0B & 45\% & 25.70 & 37.92 & 3.42 & 200.00 & 618.25 & 83.12 & 34.35 & 60.66\% \\
& 1.6B & 60\% & 24.80 & 6.12 & 0.00 & 141.07 & 293.23 & 2.34 & 10.93 & 13.52\% \\ 
\midrule
Depth-wise & 2.8B & 15\% & 33.80 & 29.42 & 69.66 & 271.43 & 1456.41 & 87.91 & 54.59 & 91.52\% \\
& 2.5B & 30\% & 29.00 & 24.77 & 28.76 & 272.86 & 1273.34 & 86.50 & 44.47 & 74.55\% \\
& 2.0B & 45\% & 23.90 & 16.85 & 3.47 & 191.43 & 867.37 & 80.09 & 31.94 & 53.54\% \\
& 1.6B & 60\% & 26.60 & 0.02 & 17.15 & 0.71 & 55.92 & 0.02 & 7.78 & 13.04\% \\
\bottomrule
\end{tabular}
}
\label{Combined Pruning Results}
\end{center}
\vspace{-10pt}
\end{table}

\xhdr{Comparison of Pruning Techniques.}
We detail the complete results in Table \ref{Combined Pruning Results} and contrast layerwise and widthwise pruning across Bunny and LLaVA in Figure \ref{fig:finetune_plot}. Without any recovery training (blue curves), widthwise pruning consistently preserves more accuracy, retaining 95\% of baseline performance on Bunny and 93\% on LLaVA at a modest 15\% compression, making it a practical choice when compute or data for recovery are scarce and only light pruning (\textless 20\%) is required.
As compression deepens, performance for both methods declines sharply. Overall, widthwise pruning better preserves the model’s structure and information flow, allowing it to keep performance with minimal adjustments, especially at lower compression ratios.
Adding recovery training reshapes the landscape (represented by the green lines). For smaller compression ratios (\textless 40\%), layerwise pruning offers a slight advantage, while widthwise pruning delivers better overall performance for larger compression ratios (\textgreater 40\%). This suggests that finetuning plays a crucial role in reconstructing inter-layer connections and reoptimizing layer components. 

\textbf{\textit{\textcolor{blue!50!black!80}{Takeaway.}}}
A widthwise pruned model can often be deployed without recovery training with a small compression ratio (\textless $20$\%). 
With recovery training, layerwise pruning shows a slight advantage at compression ratios below 30\%, while widthwise pruning performs better at higher compression ratios.

\begin{wraptable}{r}{0.5\textwidth} 
\centering
\vspace{-32pt}
\caption{Memory requirements (Mem.) and FLOPS for the Bunny and LLaVA models at various compression ratios. The models are pruned widthwise. Memory and compute reduction is significant with higher compression ratios.}
\vspace{2pt}
\resizebox{\linewidth}{!}{
\begin{tabular}{c|c|c|c|c}
\toprule
\multirow{2}{*}{Ratio} &\multicolumn{2}{c}{Bunny} &\multicolumn{2}{c}{LLaVA} \\
& Mem. (MiB) & FLOPS (T) & Mem. (MiB) & FLOPS (T) \\ 
\midrule 
0\%   & 6,167 & 4.77 & 13,546 & 9.57 \\
15\%  & 5,380 & 4.14 & 11,530 & 8.21 \\
30\%  & 4,597 & 3.50 & 9,548  & 6.89 \\
45\%  & 3,770 & 2.84 & 7,470  & 5.49 \\
60\%  & 2,992 & 2.20 & 5,435  & 4.17 \\ 
\bottomrule
\end{tabular}
}
\label{tab:model_size_widthwise}
\vspace{-20pt}
\end{wraptable}

\xhdr{From compression ratio to resource usage.} 
Table \ref{tab:model_size_widthwise} provides an overview of how different compression ratios impact memory usage and FLOPS for both the models compressed via widthwise pruning. Memory consumption refers to the allocated GPU memory, while FLOPS are measured using the Calflops codebase\footnote{Calflops codebase: https://github.com/MrYxJ/calculate-flops.pytorch}. The results demonstrate that higher compression ratios consistently lead to both memory and compute reductions. For example, at a 30\% compression ratio, we observe a memory reduction of 25\% for Bunny and 28\% for LLaVA, with a corresponding decrease in FLOPS of 27\% for both models. 
These reductions continue to scale with larger compression ratios; at a 60\% compression ratio, memory usage and FLOPS decrease by 50-60\%.
We observe similar results for layerwise pruning (see Appendix \ref{memory requirement layerwise}). This indicates that the compressions directly translate into improvements in memory efficiency and computational cost.

\subsection{Supervised Finetuning for Performance Recovery}
\begin{figure*}[t]
\vspace{-5pt}
    \centering
    \includegraphics[width=\linewidth]{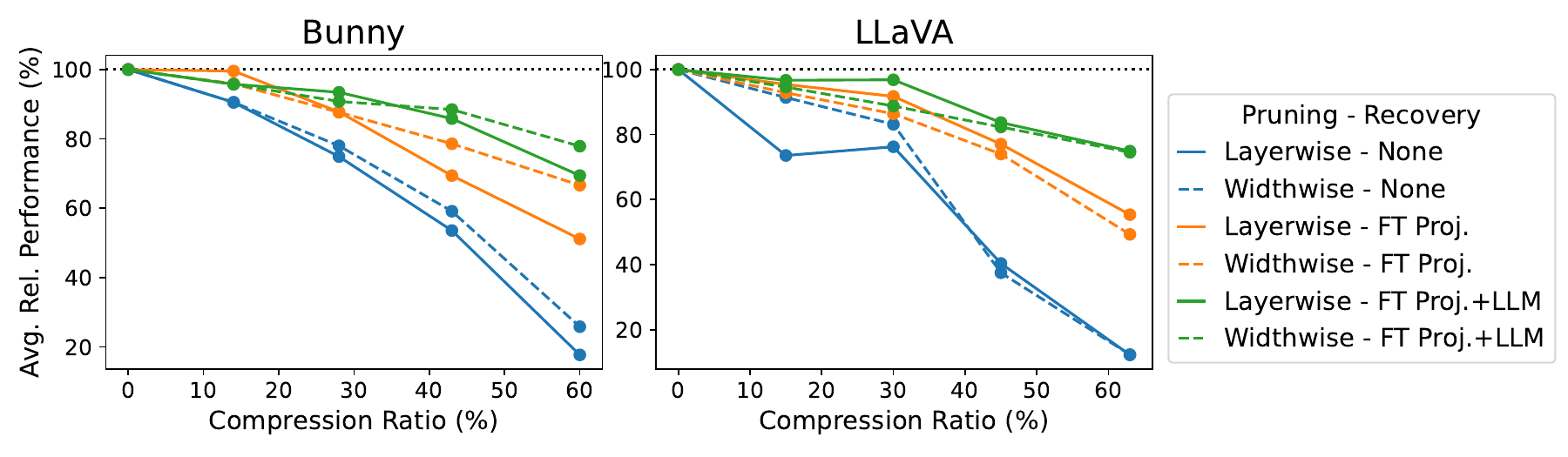}
    \vspace{-20pt}
    \caption{Comparison of pruning and finetuning strategies on two MLLMs. The plot shows the average relative performance under three scenarios: pruning only, pruning followed by finetuning the projector, and pruning followed by finetuning both the projector and the LLM. For smaller compression ratios~(\textless 20\%), finetuning only the projector effectively recovers performance. For larger compression ratios, jointly finetuning the projector and the LLM is needed for a better recovery.
    }
    \label{fig:finetune_plot}
\vspace{-18pt}
\end{figure*}

\label{Supervised Finetuning for Performance Recovery After Pruning}
Compressing LLMs can degrade their language modeling capabilities. More critically, the impact of pruning LLM decoders (within MLLMs) on visual understanding and the alignment between vision and language remains largely unexplored. To investigate these effects, we experiment with two approaches: (1) finetuning only the multimodal projector and (2) jointly finetuning both the projector and the LLM. This enables us to pinpoint the source of performance degradation and assess the extent to which each component contributes to the model's overall effectiveness.
Following the previous research \cite{karamcheti2024prismatic}, which shows that training the vision encoder may degrade overall model performance, we keep the vision encoder frozen in both setups. To facilitate fast recovery, we employ the low-rank approximation, LoRA \cite{hu2021lora}, while finetuning the LLM.

\xhdr{Finetuning the multimodal projector.} 
As shown in Figure \ref{fig:finetune_plot}~(orange lines), finetuning the multimodal projector significantly restores performance. At lower compression ratios (\textless $20$\%), finetuning only the projector achieves results comparable to jointly finetuning the LLM. For both Bunny and LLaVA, finetuning the projector retains at least 95\% of the performance at a compression ratio of 15\%.  As the compression ratio increases, the loss of language modeling ability becomes more pronounced, making projector-only finetuning insufficient to fully recover the model’s performance. Nevertheless, even at a compression ratio of 60\%, finetuning the multimodal projector can still recover 60 to 80\% of the performance by realigning the vision and language inputs.
This indicates that pruning specific LLM structures in the MLLM can both impair the language modeling ability and introduce modality misalignment, thereby hindering the model's ability to comprehend visual inputs.

\xhdr{Finetuning both the projector and the LLM.}
While a significant portion of the recovered performance is attributed to realigning the visual and textual inputs, we observe consistent gains from additionally finetuning the pruned LLM~(green lines in Figure~\ref{fig:finetune_plot}), especially at higher compression ratios (\textgreater 40\%). This indicates that the pruned model not only suffers from modality misalignment but also experiences a decline in its language modeling capabilities. We can partly restore these lost capabilities by finetuning the LLM. At a compression ratio of 40\%, finetuning both the projector and the LLM restores more than 80\% of the original model’s performance. Even at a compression ratio of 60\%, finetuning recovers close to 80\% of the model’s original performance.

\textbf{\textit{\textcolor{blue!50!black!80}{Takeaway.}}}
When a small compression ratio of around 15\% is required, finetuning the multimodal projector alone is typically sufficient to recover most of the model’s performance. For higher compression ratios (\textgreater 40\%), incorporating finetuning of the LLM yields additional performance improvements.

\begin{figure*}[t]
    \centering
    \includegraphics[width=\linewidth]{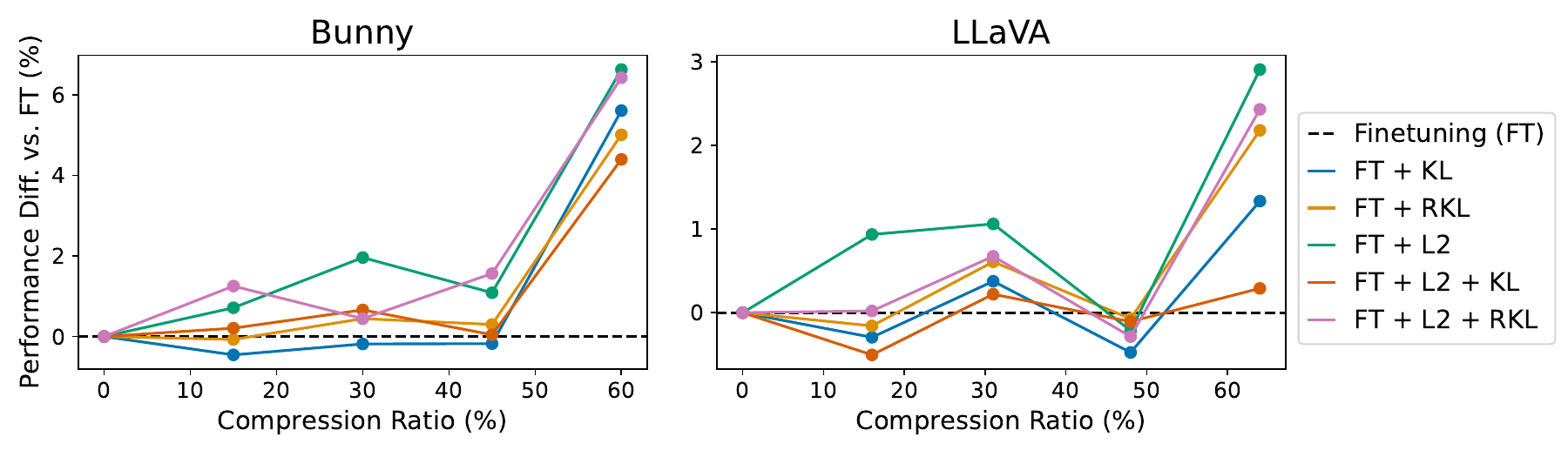}    
    \vspace{-20pt}
    \caption{Comparison of different distillation recovery strategies (KL loss, RKL loss, L2 loss, and their combinations) for Bunny and LlaVA models pruned with widthwise pruning. The plot shows the relative performance improvement of each strategy over standard finetuning across various compression ratios. The results demonstrate that distillation helps recover more performance than finetuning alone, with the L2 loss component consistently leading to the largest performance gains.
    }
    \label{fig:distillation_llmpruner_plot}
\vspace{-10pt}
\end{figure*}

\subsection{Knowledge Distillation for Performance Recovery After Pruning}
\label{kd_for_recovery}

To compare and analyze the effectiveness of FT and KD, we present the recovered results for the layerwise pruned Bunny model. We compare a logit-based approach (RKL) and a hidden state matching strategy (L2), with and without a finetuning loss component in Table \ref{tab:dist_w_ft}.
At a light 15 \% compression, pure distillation already recovers 95 \% of the baseline accuracy, with FT alone performing similarly (96.3\%). However, as the compression ratio increases, these KD-only variants become unstable: at 60 \% compression, L2 degrades to 47.6 \% and RKL to just 12.6 \%. Crucially, coupling FT with KD not only prevents this collapse but delivers the best results across the board, providing consistent gains of 3 to 23 percentage points.

\begin{wraptable}{b} {0.5\textwidth} 
\vspace{-32pt}
\caption{Comparison of distillation strategies with and without finetuning for the Bunny model compressed via layerwise pruning. We show the performance ratio between the compressed model and the original model. Finetuning helps stabilize performance and prevents \textit{model collapse}, especially at higher compression ratios.}
\vspace{2pt}
\centering
\resizebox{0.5\textwidth}{!}{ 
\begin{tabular}{c|c|cc|cc}
\toprule
 & \multicolumn{5}{c}{Bunny} \\
Ratio & FT & L2 & L2+FT & RKL & RKL+FT \\
\midrule
15\% & 96.30\% & 95.51\% & \textbf{99.59\%} & 96.88\% & \textbf{98.70\%} \\
30\% & 94.33\% & 88.13\% & \textbf{95.03\%} & 92.21\% & \textbf{93.81\%} \\
45\% & 86.70\% & \textit{56.96\%} & \textbf{90.19\%} & 82.57\% & \textbf{88.50\%} \\
60\% & 69.38\% & \textit{47.61\%} & \textbf{72.62\%} & \textit{12.61\%} & \textbf{69.85\%} \\
\bottomrule
\end{tabular}
}
\label{tab:dist_w_ft}
\vspace{-20pt}
\end{wraptable}

The pattern underscores a clear message: while KD can partially restore performance after pruning, its reliability declines at high compression; adding FT supplies the hard-label anchor that stabilizes learning, while the soft guidance of distillation offers complementary structural cues, making the FT + KD combination the most dependable strategy across the entire compression spectrum (Full results in Appendix~\ref{Knowledge distillation with and without finetuning}).

Figure \ref{fig:distillation_llmpruner_plot} compares various distillation strategies based on their relative improvement over finetuning alone when widthwise pruning is applied (see Appendix \ref{kd+ft} for further results on layerwise pruning). Our results indicate that applying the L2 loss to align the hidden states of the student and teacher in the final layer yields the best performance, or at least matches other methods. Unlike logit-based approaches, which require the student to replicate the teacher’s output distribution, the L2 loss method enables the student to directly capture the teacher’s feature representations, leading to enhanced performance.
Additionally, we observe that RKL generally outperforms KL across most compression ratios, a result consistent with the findings of \cite{gu2024minillmknowledgedistillationlarge}.

\textbf{\textit{\textcolor{blue!50!black!80}{{Takeaway.}}}}
Knowledge distillation, particularly when combined with finetuning and using L2 loss to map the intermediate states, delivers the most effective performance recovery after pruning across all compression ratios.

\subsection{Data Efficient Recovery} 
\label{Experiments_few_data}
\begin{figure*}[t]
    \centering
    \includegraphics[width=\linewidth]{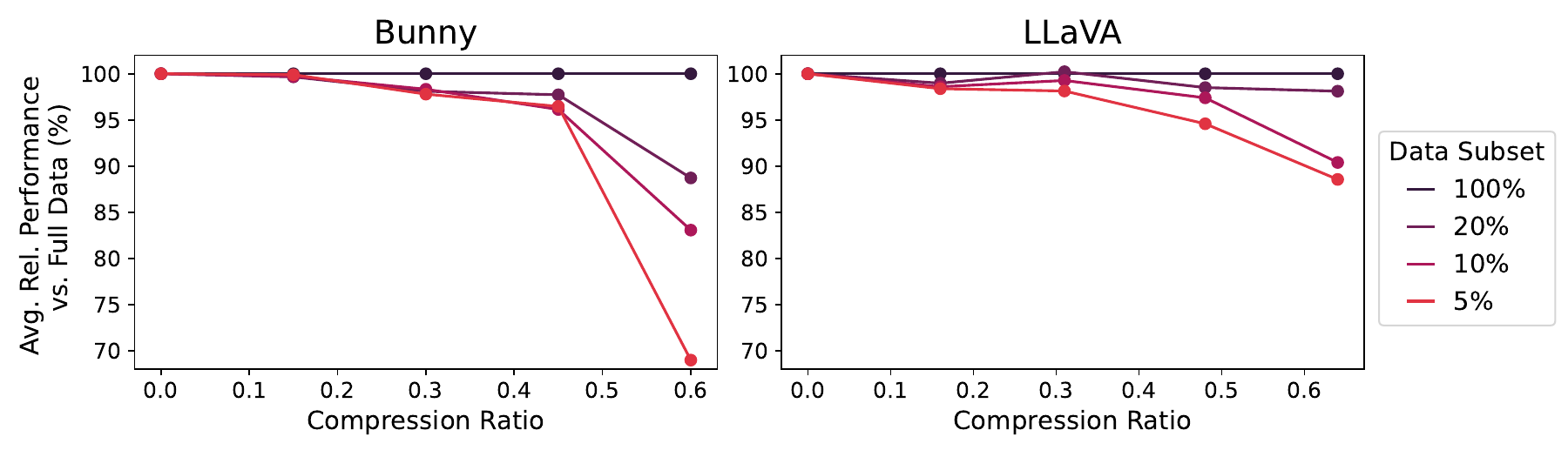}   
    \vspace{-20pt}
    \caption{Comparison of recovery performance using different percentages of training data (100\%, 20\%, 10\%, and 5\%) for finetuning and distillation after pruning across Bunny and Llava models. For smaller compression ratios, even a small percentage of the training data (as low as 5\%) is sufficient to recover most of the original performance. However, as the compression ratio increases, more training data is required to achieve higher recovery performance.
    }
    \label{fig:fewdata_plot}
\vspace{-8pt}
\end{figure*}

Figure \ref{fig:fewdata_plot} shows the models' performance after recovery training with different portions of the original dataset relative to training with the full 100\%. Both models undergo widthwise pruning and recovery training, incorporating RKL and L2 loss functions. Remarkably, for compression ratios below 50\%, using just 5\% of the original data is sufficient to achieve over 95\% of the performance compared to using the full dataset. However, as the compression ratio increases, the amount of data required for effective recovery training also grows. For a compression ratio of 60\%, the relative performance drops below 90\% for LLaVA, and further diminishes to below 70\% for Bunny. Nevertheless, using only a small portion of the training data appears to be a valid option, significantly lowering the required time and cost for compressing and finetuning MLLMs. 

\textbf{\textit{\textcolor{blue!50!black!80}{{Takeaway.}}}}
With a compression ratio smaller than 50\%, using just 5\% of the dataset is enough to achieve performance comparable to full data training. However, for compression ratios greater than 50\%, full data training becomes necessary to recover performance effectively.

\subsection{Key Insights for Model Compression}
\label{Model compression results following our best practices}
Based on the empirical results from the previous section, we outline the following suggested practices for compressing MLLMs:
\vspace{3pt}
\begin{tcolorbox}[colback=gray!10, colframe=blue!20!gray, rounded corners=all, before skip=0pt,
    left=0pt]
    \setlength{\leftmargini}{10pt}
    \begin{itemize}
        \item \textbf{Widthwise pruning is more effective in low-resource settings}, yielding an efficient model even without the need for recovery training.
        \item \textbf{With recovery training}, layerwise pruning excels for smaller compression ratios (\textless 40\%), while widthwise pruning performs better at higher ratios (\textgreater 40\%).
        \item \textbf{For small compression ratios (\textless 20\%)}, finetuning just the multimodal projector is often sufficient to restore performance, with minimal impact from pruning.
        \item \textbf{For recovery training}, combining finetuning with knowledge distillation of the intermediate representations using L2 loss consistently achieves the highest performance across all compression ratios.
        \item \textbf{Data efficiency} can be significantly boosted, requiring only 5\% of the original data to match full-data training results, though full datasets are still needed for high compression ratios. 
    \end{itemize}
\end{tcolorbox} 
In Appendix \ref{Detailed model compression results following our best practices} and \ref{Qualitative results of the compressed models} we report quantitative and qualitative results for the compressed models. To assess the generalizability of our best practices, we extended our experiments to Mini-InternVL-Chat-4B-V1-5 \cite{chen2024far}. We provide complete results in Appendix \ref{generalizability_best_practices}, which show the generalizability of our insights.

\section{Discussion}

\subsection{How does pruning LLM impact multi-modal capability?}
\label{How does pruning LLM impact modality alignment?}
Since LLMs comprise the majority of parameters in MLLMs, reducing their size can substantially reduce the overall model's size. However, pruning an LLM presents a dual challenge: it only degrades language modeling capabilities but also disrupts the alignment between modalities, impairing the model’s ability to interpret and reason about visual inputs effectively.
Our analysis shows that for pruning ratios below 10\%, the model retains most of its multimodal functionality. With moderate pruning (up to 15\%), modality alignment can still be restored by post-training the multimodal projector. However, beyond this threshold, the degradation of language modeling becomes more pronounced. At higher compression levels, finetuning the projector alone is insufficient to recover performance, making joint training of the LLM necessary to maintain functionality.

\subsection{Comparison and combination with quantization}
Integrating quantization into our framework can further optimize inference time and memory efficiency. 
In this section, we provide a comparative analysis of structured pruning and quantization, highlighting their complementary strengths when combined. As a representative quantization method, we employ LLM.int8() (\cite{dettmers2022gpt3}). As shown in Table \ref{tab:quantization}, LLM.int8() reduces memory usage by 44.5\% in the original uncompressed model, while incurring only a minor performance loss of 0.43 percentage points. However, this comes at the cost of a fourfold increase in latency. For LLaVA-6B and LLaVA-5B, combining pruning with quantization offers a well-balanced trade-off between memory efficiency and computational latency.
\begin{table}[h]

\caption{This table compares pruning and quantization applied to LLaVA-v1.5-7b, evaluating their effect on memory usage, average performance across benchmarks, and inference latency. Quantization significantly reduces memory consumption but increases latency, while pruning with recovery maintains a balance between efficiency and performance. Combining both techniques mitigates quantization overhead while preserving compression benefits.}
\centering
\resizebox{0.7\linewidth}{!}{
\begin{tabular}{lccccccccccc}
\toprule
Model & Quantization & Mem (GiB) & \ \  \ Ratio \  \ & \  \ \ Avg  \ \ \ \ & Latency (ms) \\ 
\midrule
LLaVA-7B &-& 13.5 & 0\% & 62.28 & 105 ± 1.5 \\ 

LLaVA--7B & \checkmark & 7.5 & 0\% & 61.85 & 398 ± 1.3 \\ 
\midrule

LLaVA-6B & -&11.6 & 15\% & 61.22 & 95 ± 8.1 \\ 

LLaVA-6B & \checkmark & 6.5 & 15\% & 60.82 & 125 ± 0.9 \\ 
\midrule

LLaVA-5B &-& 9.5 & 30\% & 60.96 & 80.7 ± 0.6 \\ 

LLaVA-5B & \checkmark & 5.4 & 30\% & 59.63 & 108 ± 5.85 \\ 
\toprule
\end{tabular}}
\label{tab:quantization}
\vspace{-15pt}
\end{table}

\subsection{Limitation and future work}
\label{Limitation and future work}
Our experiments demonstrate the effectiveness of structural pruning with recovery training at moderate compression ratios (up to 30\%). 
However, beyond this threshold, performance loss becomes increasingly difficult to recover, suggesting that for applications requiring more aggressive compression, the extreme pruning of a large model is not a viable approach.
Due to computational constraints, this work focuses on two pruning techniques applied to three different models. Future work could extend these findings to include a broader range of pruning techniques and models, further refining these strategies. 
\section{Conclusion}
\label{conclusion}
We systematically evaluated two structural pruning schemes---widthwise and layerwise---on LLaVA-7B, Bunny-3B, and InternVL, and paired them with lightweight recovery through supervised finetuning and knowledge distillation. 
From these experiments, we distilled a decision chart that guides practitioners in choosing the pruning route and recovery budget for different target compression ratios.
Our findings provide a concrete path to fit MLLMs within strict memory, compute, or energy budgets without surrendering performance.
\clearpage

\bibliographystyle{splncs04}
\bibliography{main}
\clearpage
\appendix
\section{Appendix}

\subsection{Model architecture of the LLaVA and Bunny models used in the main experiments}
\label{Model architecture of LLaVA and Bunny}
Table \ref{model architecture} outlines the architectures of the Bunny and LLaVA models. 
LLaVA is built upon Vicuna-v1.5 \cite{vicuna2023} with 6.7 billion parameters, and Bunny is based upon Phi-2 \cite{javaheripi2023phi} with 2.8 billion parameters. LLaVA-v1.5-7B employs CLIP-ViT-L \cite{radford2021learning} as the vision encoder and Vicuna-v1.5 \cite{vicuna2023} as the language decoder, while Bunny-v1.0-3B utilizes SigLIP-SO \cite{zhai2023sigmoid} as the vision encoder and Phi-2 \cite{javaheripi2023phi} as the language decoder. Both models leverage MLP layers to align the vision and language modalities.

\begin{table}[h]
\vspace{-10pt}
\caption{Architecture details of the uncompressed models. We present the number of parameters, along with the vision encoder, multimodal projector and the language decoder of the models included in our study.}
\centering
\resizebox{\linewidth}{!}{
\begin{tabular}{ccccc}
\toprule
\multicolumn{1}{c}{\bf Model}
&\multicolumn{1}{c}{\bf Parameters}
&\multicolumn{1}{c}{\bf Vision Encoder}
&\multicolumn{1}{c}{\bf Multimodal Projector}
&\multicolumn{1}{c}{\bf Language Decoder}
\\ \midrule 
LLaVA-v1.5-7B & 7.0B & CLIP-ViT-L~(0.3B) & mlp2x-gelu~(0.01B) & Vicuna-v1.5~(6.7B) \\
Bunny-v1.0-3B & 3.2B & SigLIP-SO~(0.4B) & mlp2x-gelu~(0.02B) & Phi-2~(2.8B)
\\ \bottomrule
\end{tabular}
}
\label{model architecture}
\vspace{-10pt}
\end{table}

\subsection{Implementation details of hidden state matching}
\label{L2 layer comparison}
\begin{table}[h]
\vspace{-20pt}
\centering
\caption{Results for recovering widthwise pruned Bunny with hidden state mapping. We compare the relative performance for mapping the last layer (layer-1), the last two layers (layer-1,2), and the last three layers (layer-1,2,3). By only mapping the last LLM layer the best performance is achieved. }
\begin{center}
\begin{tabular}{rrrr}
\toprule
\multicolumn{1}{c}{\bf Ratio}  
&\multicolumn{1}{c}{\bf Layer-1}  
&\multicolumn{1}{c}{\bf Layer-1,2}
&\multicolumn{1}{c}{\bf Layer-1,2,3}
\\ \midrule 
12.8\% & 95.34\% & 95.17\% & 96.25\% \\
25.5\% & 91.02\% & 90.48\% & 90.97\% \\
39.0\% & 87.08\% & 86.12\% & 84.84\% \\
51.8\% & 75.25\% & 72.56\% & 72.87\% \\
\bottomrule
\end{tabular}
\label{Bunny_L2_layer_compare}
\end{center}
\vspace{-20pt}
\end{table}

To determine which LLM layers' hidden states to map between the pruned and unpruned models, we explore three options: matching the last layer, the last two, and the last three layers. Table \ref{Bunny_L2_layer_compare} shows that matching only the last layer's hidden state yields the best performance.

\subsection{Detailed results on the efficiency of the pruned models}
\label{memory requirement layerwise}

\begin{table}[h]
  \centering
  \caption{Inference-time memory and compute cost for \textbf{layer-wise pruned} Bunny-3B and LLaVA-7B at different compression ratios. Each measurement assumes one image and a 50-token prompt.}
  \label{tab:model_size_layerwise}
  \resizebox{0.5 \linewidth}{!}{
  \begin{tabular}{c|rr|rr}
    \toprule
    \multirow{2}{*}{Compression} & \multicolumn{2}{c|}{\textbf{Bunny}} & \multicolumn{2}{c}{\textbf{LLaVA}} \\
    & Mem.\,(MiB) & FLOPs\,(T) & Mem.\,(MiB) & FLOPs\,(T) \\
    \midrule
    0 \%  & 6\,167 & 4.77 & 13\,546 & 9.57 \\
    15 \% & 5\,411 & 4.16 & 11\,604 & 8.03 \\
    30 \% & 4\,659 & 3.56 & 9\,664  & 6.92 \\
    45 \% & 3\,907 & 2.95 & 7\,724  & 5.55 \\
    60 \% & 3\,006 & 2.22 & 5\,496  & 3.90 \\
    \bottomrule
  \end{tabular}}
\end{table}

For the models pruned by layerwise method, we also assess their memory consumption as FLOPs. Memory consumption refers to the allocated GPU memory, while FLOPS are measured using the Calflops codebase. The results in Table \ref{tab:model_size_layerwise} show the same trend as widthwise pruning, indicating that the achieved compressions directly translate into improvements in memory efficiency and computational cost for both widthwise and layerwise pruning.

\subsection{Is SFT needed for KD?}
\label{Knowledge distillation with and without finetuning}
\begin{figure*}[h]
    \centering
    \includegraphics[width=\linewidth]{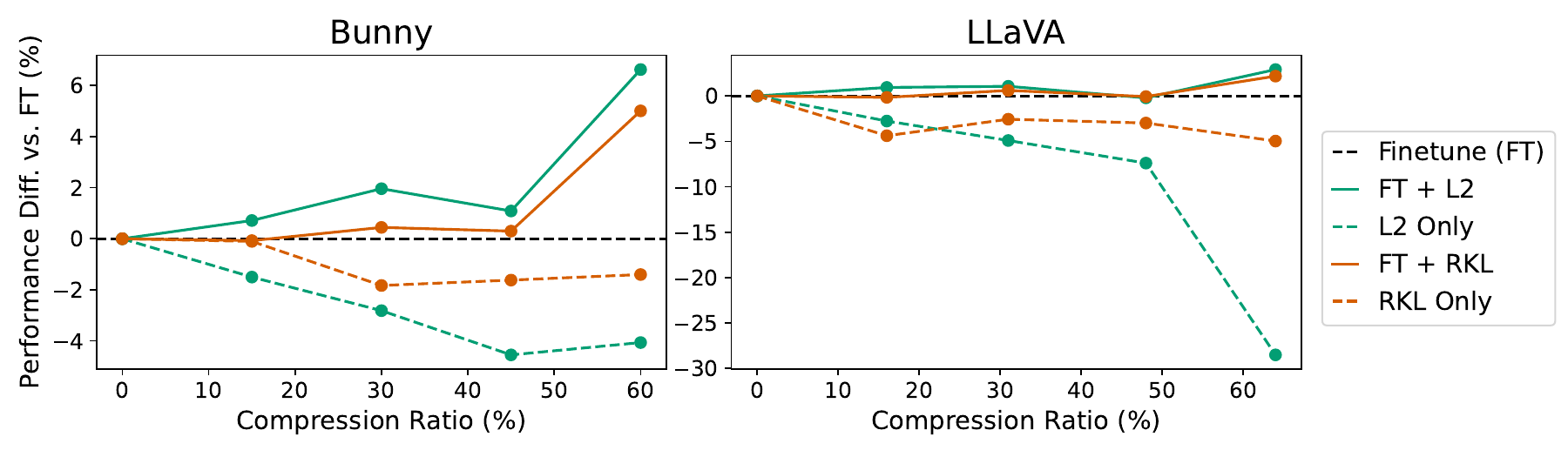}    
    \caption{Comparison of L2 and RKL distillation strategies with and without additional fine-tuning loss for Bunny and LLaVA models compressed by \textbf{widthwise} pruning. 
    The plot shows performance differences relative to standard fine-tuning across varying compression ratios. 
    }
    \label{fig:distillation_2_llmpruner_plot}
\end{figure*}

\begin{figure*}[h]
    \centering
    \includegraphics[width=\linewidth]{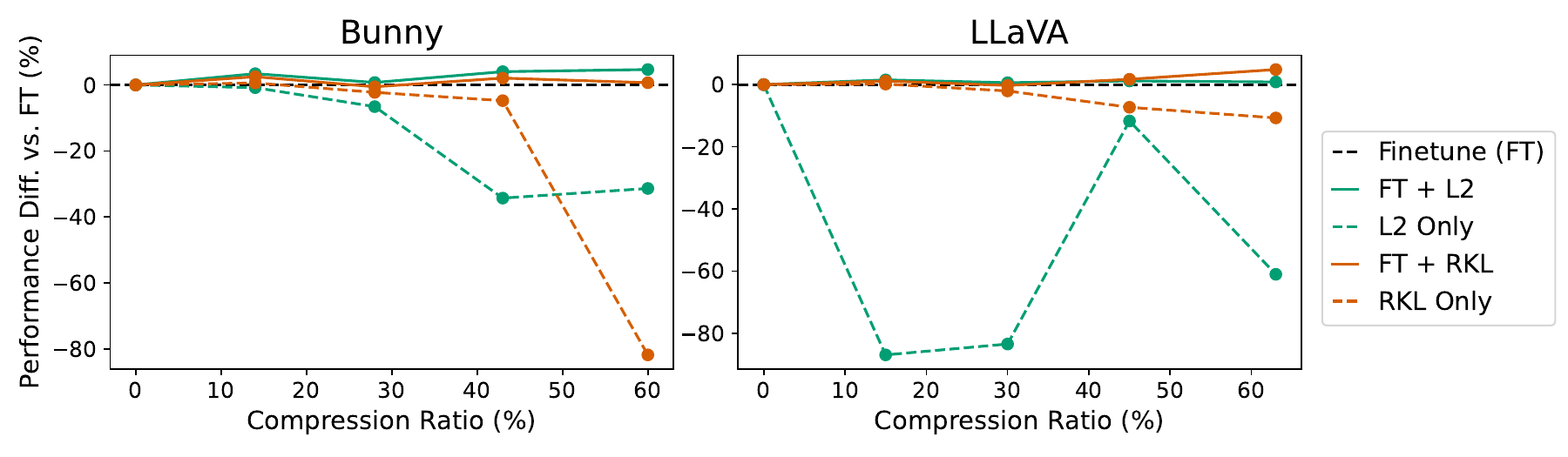}    
    \caption{Comparison of L2 and RKL distillation strategies with and without additional fine-tuning loss for Bunny and LLaV models compressed by \textbf{layerwise} pruning. The plot shows performance differences relative to standard fine-tuning across varying compression ratios. 
    }
    \label{fig:distillation_2_shortgpt_plot}
\end{figure*}

Figure \ref{fig:distillation_2_llmpruner_plot} and Figure \ref{fig:distillation_2_shortgpt_plot} compare logits-based knowledge distillation (represented by RKL) and hidden state matching-based knowledge distillation (represented by L2 loss), with and without supervised finetuning, following widthwise and layerwise pruning, respectively. While knowledge distillation alone helps in recovering performance post-pruning, it remains less effective than supervised finetuning. However, when combined with supervised finetuning, it results in superior performance.

\subsection{Which KD strategy to use after layerwise pruning?}
\label{kd+ft}
\begin{figure*}[h]
    \centering
    \includegraphics[width=\linewidth]{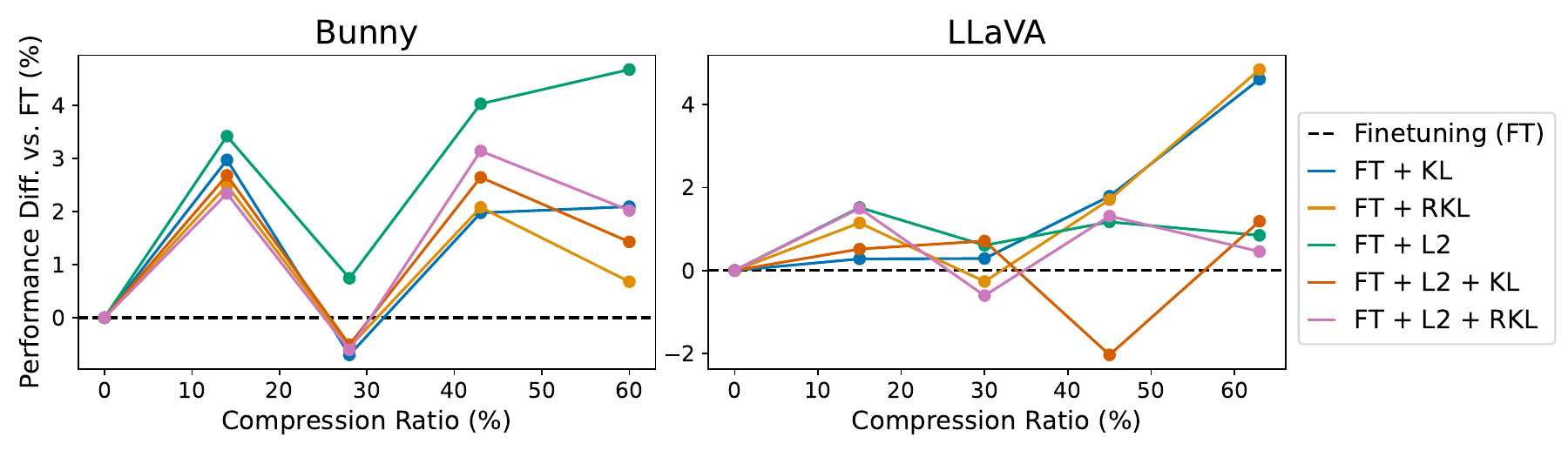}  
    \caption{Comparison of different distillation recovery strategies (KL loss, RKL loss, L2 loss, and their combinations) for Bunny and LlaVA models pruned with layerwise pruning. The plot shows the relative performance improvement of each strategy over standard fine-tuning across various compression ratios. The results demonstrate that distillation helps recover more performance than fine-tuning alone, with the L2 loss component consistently leading to the largest performance gains, particularly at higher compression ratios.}
    \label{fig:distillation_shortgpt_plot}
\end{figure*}

Figure \ref{fig:distillation_shortgpt_plot} compares various distillation strategies based on their relative improvement over finetuning alone after layerwise pruning. Similar to the results after widthwise pruning, applying hidden states matching yields the best performance, or at least matches other methods. The trend that RKL generally outperforms KL is also observed here.

\subsection{Detailed model compression results following our insights}
\label{Detailed model compression results following our best practices}
\begin{table}[t!]
\begin{center}
\caption{Performance of the best compressed models. The size is the number of total parameters of the model, while the ratio, short for compression ratio, indicates the proportion of remaining LLM parameters compared to the pre-pruning state. When the compression ratio (Ratio) is below 40\%, we apply depthwise pruning. For ratios above 40\%, we use widthwise pruning. During the recovery phase, we employ supervised finetuning combined with L2 loss to match the hidden states. For both Bunny and LLaVA, 95\% performance is retained if the compression ratio is smaller than 40\%.}
\vspace{2pt}
\resizebox{\linewidth}{!}{
\begin{tabular}{rrrrrrrrrrr}
\toprule

\multicolumn{1}{c}{\bf Method}
&\multicolumn{1}{c}{\bf Size}  
&\multicolumn{1}{c}{\bf Ratio}  
&\multicolumn{1}{c}{\bf MMMU}
&\multicolumn{1}{c}{\bf GQA}
&\multicolumn{1}{c}{\bf SQA}
&\multicolumn{1}{c}{\bf MME-C}
&\multicolumn{1}{c}{\bf MME-P}
&\multicolumn{1}{c}{\bf POPE}
&\multicolumn{1}{c}{\bf AVG}
&\multicolumn{1}{c}{\bf AVG-\%}

\\ \midrule 
\multicolumn{11}{c}{\bf Bunny-v1.0-3B}
\\ \midrule[0.00005\arrayrulewidth]
 &3.2B & 0\% & 34.10 & 54.72 & 70.70 & 289.30 & 1487.71 & 87.82 & 59.65 & 100.00\% \\
Depth+FT+L2&2.8B & 15\% & 33.00 & 54.56 & 70.00 & 304.29 & 1457.06 & 87.97 & 59.40 & 99.59\% \\
Depth+FT+L2&2.5B & 30\% & 32.30 & 53.08 & 68.12 & 252.50 & 1349.91 & 87.53 & 56.68 & 95.03\% \\
Width+FT+L2&2.0B & 45\%  & 29.10 & 52.31 & 63.06  & 244.64  & 1281.66 & 87.09 & 54.37 & 91.15\%  \\
Width+FT+L2&1.6B & 60\% & 28.10 & 48.72 & 53.20  & 216.07  & 1115.33 & 86.73 & 49.92 & 83.69\%  \\
\midrule 
\multicolumn{11}{c}{\bf LLaVA-v1.5-7B}
\\ \midrule[0.00005\arrayrulewidth]
&7.0B & 0\% & 35.10 & 61.98 & 68.67 & 363.21 & 1511.33 & 86.99 & 62.28 & 100.00\% \\
Depth+FT+L2&6.3B & 15\% & 36.40 & 61.20 & 68.42 & 337.86 & 1442.35 & 86.94 & 61.22 & 98.29\% \\
Depth+FT+L2&5.5B & 30\% & 36.00 & 60.34 & 68.82 & 318.57 & 1496.60 & 85.98 & 60.96 & 97.88\% \\
Width+FT+L2& 3.8B & 45\% & 30.80 & 57.74 & 52.90 & 215.00 & 1191.17 & 85.74 & 52.27 & 83.92\% \\
Width+FT+L2& 2.8B & 60\% & 27.70 & 52.32 & 46.26 & 211.79 & 1085.97 & 84.06 & 48.52 & 77.90\% \\
\bottomrule
\end{tabular}
}
\label{tab:best_compressed_model}
\end{center}
\vspace{-10pt}
\end{table}

This section provides detailed numerical results of the model performance following our findings and insights. To illustrate the performance at different compression ratios, Table \ref{tab:best_compressed_model} offers a detailed comparison of results for both Bunny and LLaVA across various multimodal benchmarks. The results show that, with compression ratios below 30\%, Bunny retains over 95\% of its original performance, while LLaVA maintains more than 97\%. Even at higher compression ratios, up to 60\%, our method preserves an average performance of 83\% for Bunny and 78\% for LLaVA. These findings underscore the feasibility of compressing MLLMs without incurring significant performance degradation.

\subsection{Qualitative results of the compressed models}
\label{Qualitative results of the compressed models}
In this section. We present some qualitative results of the compressed models. Table \ref{tab:qulitative_analysis_llava_models} presents the qualitative evaluation results of the compressed LLaVA models. Despite undergoing compression, these models exhibit a remarkable capacity for understanding and processing visual inputs with high accuracy. They effectively analyze images and generate rich, detailed textual descriptions. This demonstrates that compression does not significantly compromise their ability to comprehend complex visual information. Instead, the models maintain strong performance, producing coherent and contextually relevant outputs.

\subsection{Generalizability of our insights}
\label{generalizability_best_practices}
We extended our experiments to Mini-InternVL-Chat-4B-V1-5, which comes from the recent InternVL model family\cite{chen2024far}. It comprises of  InternViT-300M-448px as vision encoder, and Phi-3-mini-128k-instruct as the LLM. As shown in Table \ref{Internvl_combine}, widthwise pruning outperforms layerwise pruning without recovery training, preserving 97.4\% of the original performance at 15\% compression compared to 96.7\% for layerwise pruning. This reinforces widthwise pruning as the preferred strategy in low-resource scenarios.
We further examined the impact of recovery training, including finetuning the multimodal projector and the LLM, as well as the importance of incorporating knowledge distillation. Table \ref{Internvl_combine} shows that at a 15\% compression, projector-only finetuning restores 96.9\% of the original performance, while jointly finetuning the projector and LLM improves recovery to 97.8\%. At 30\% compression, these numbers drop to 75.1\% and 86.6\%, respectively. Supervised finetuning with hidden state based distillation consistently yields the best results, recovering 98.2\% at 15\% compression and 87.2\% at 30\%. These findings confirm our insights generalize well across architectures, ensuring robustness and broader applicability.
\begin{table}[h!]
\centering
\caption{Performance InternVL-Chat-4B. Comparison of different pruning methods, recovery training only multimodal projector (mm) and large language model on Mini-InternVL-Chat-4B-V1-5 after layerwise pruning with different recovery strategies, i.e., supervised finetuning (SFT) and knowledge distillation (KD) on intermediate representations.}
\vspace{2pt}
\resizebox{0.7\linewidth}{!}{
\begin{tabular}{rrrrrrrr}
\toprule
\textbf{Size} & \textbf{Ratio} & \textbf{Pruning} & \textbf{$\text{SFT}_\text{mm}$} &  \textbf{$\text{SFT}_\text{all}$} & \textbf{$\text{KD}_\text{all}$} & \textbf{AVG} & \textbf{AVG-\%} \\ \midrule
4B &-&-&-&-&-& 72.56 & 100\% \\ 
\midrule
3.5B & 15\% & \text{Layerwise} &-&-&-&70.15 & 96.68\% \\ 
3.5B & 15\% & \text{Widthwise} &-&-&-&70.70 & 97.44\% \\ 
\midrule
3.5B & 15\%& \text{Layerwise} & \checkmark &\checkmark&- & 70.15 & 96.68\% \\
3.5B & 15\%& \text{Layerwise} &- &\checkmark&- & 70.96 & 97.80\% \\
3.5B & 15\%& \text{Layerwise} &- &\checkmark&\checkmark& 71.23 & 98.16\% \\
3B & 30\% & \text{Layerwise} & \checkmark &\checkmark&- & 43.94 & 60.56\% \\
3B & 30\%& \text{Layerwise} &- &\checkmark&- & 62.86 & 86.64\% \\
3B & 30\% & \text{Layerwise} &- &\checkmark&\checkmark& 63.29 & 87.23\% \\
\bottomrule
\end{tabular}
}
\label{Internvl_combine}
\end{table}

\begin{table}[b]
\begin{center}
\begin{tabular}{p{2.5cm} p{10cm}}  
\toprule
\multicolumn{1}{c}{\bf Model} 
& \multicolumn{1}{c}{\bf Response}
\\ \midrule 
&\includegraphics[width=0.8\linewidth]{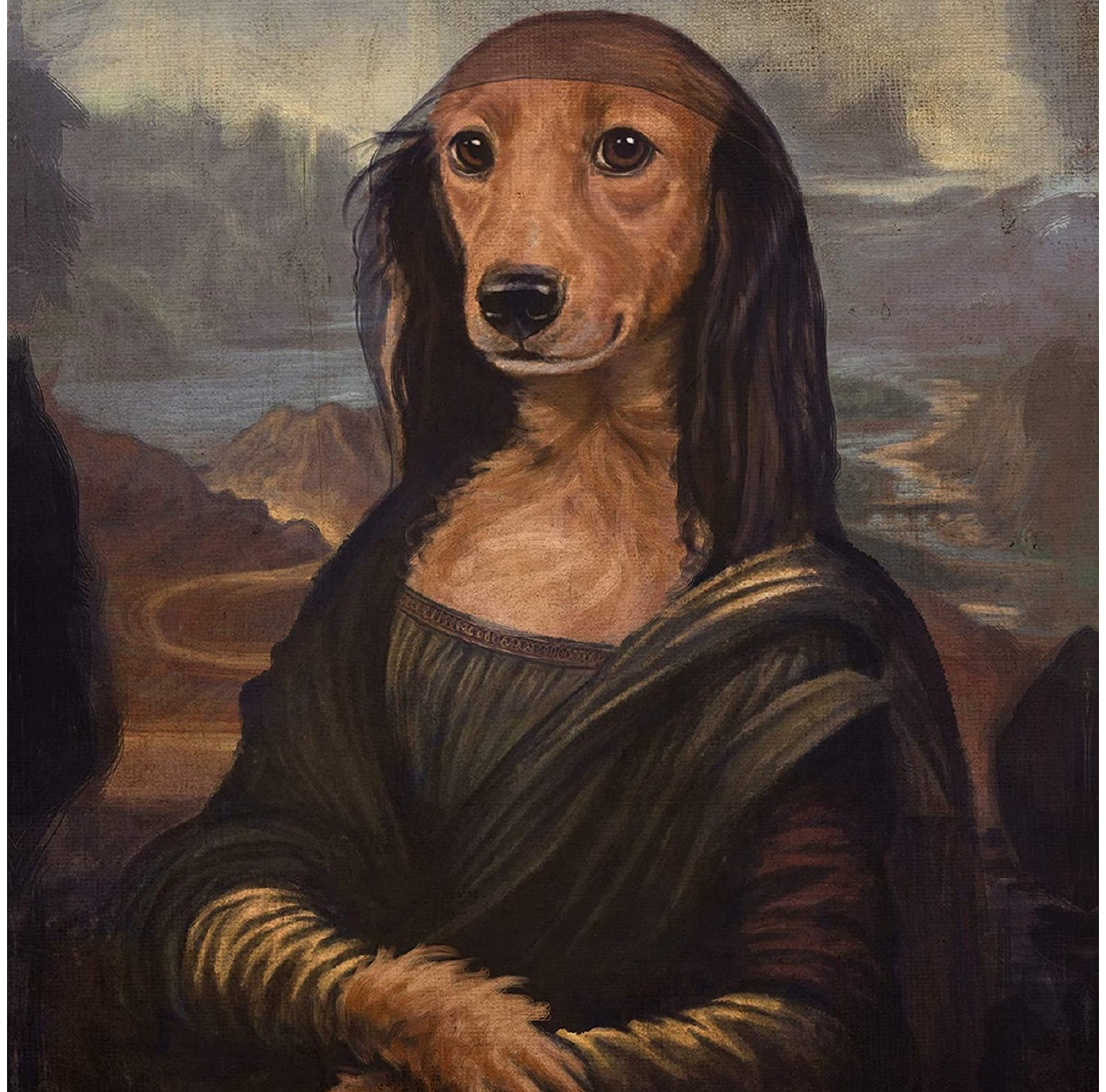} \\
&{\bf Prompt}: A chat between a curious user and an artificial intelligence assistant. The assistant gives helpful, detailed, and polite answers to the user's questions. USER: $<image>$ What is funny in the image? \\
\midrule 
LLaVA-6.3B & 
The image is funny because it features a dog wearing a Renaissance-style dress and a bandana, posing as if it's posing for a painting. This is humorous because it is a playful and unconventional representation of a dog, which is not something you would typically see in a painting or a Renaissance-style dress. The dog's pose and attire create a sense of humor and absurdity, making the image entertaining and amusing to viewers. \\
\midrule 
LLaVA-5.5B & 
The image is funny because it features a dog dressed up in a human-like attire, wearing a long black dress and a hat. This unusual and amusing scene is not something you would typically see in real life, as dogs don't wear clothes or accessories like humans. The dog's expression and the way it is posing for the picture further contribute to the comic effect of the image. The image captures the viewer's attention and sparks a sense of humor due to the unexpected and imaginative representation of a dog dressed like a human. \\
\midrule 
LLaVA-3.8B & 
The image is funny because it features a dog dressed in a human-like dress, posing in a pose as if it were a person. The dog appears to be dressed in a dress, which is a humorous and unexpected scene, as dogs are not typically dressed like humans. The dog's pose and the overall scene create a sense of humor and playfulness, making the image a delightful and entertaining piece. \\

\bottomrule
\end{tabular}
\end{center}
\caption{Qualitative analysis of compressed LLaVA models with respect to the image. We feed the compressed LLaVA models the image paired with a prompt to generate a response. The compressed LLaVA models are able to understand the visual inputs and output reasonable texts.}
\label{tab:qulitative_analysis_llava_models}
\end{table}

\end{document}